# A network analysis of decision strategies of human experts in steel manufacturing

Daniel Christopher Merten[a] , Prof. Dr. Marc-Thorsten Hütt[b] , Prof. Dr. Yilmaz Uygun[a]

[a]Department of Mathematics & Logistics, Jacobs University, Bremen, Germany

[b]Department of Life Sciences & Chemistry, Jacobs University, Bremen, Germany

# A network analysis of decision strategies of human experts in steel manufacturing

**Steel production scheduling is typically accomplished by human expert planners. Hence, instead of fully automated scheduling systems steel manufacturers prefer auxiliary recommendation algorithms. Through the suggestion of suitable orders, these algorithms assist human expert planners who are tasked with the selection and scheduling of production orders. However, it is hard to estimate, what degree of complexity these algorithms should have as steel campaign planning lacks precise rule-based procedures; in fact, it requires extensive domain knowledge as well as intuition that can only be acquired by years of business experience. Here, instead of developing new algorithms or improving older ones, we introduce a shuffling-aided network method to assess the complexity of the selection patterns established by a human expert. This technique allows us to formalize and represent the tacit knowledge that enters the campaign planning. As a result of the network analysis, we have discovered that the choice of production orders is primarily determined by the orders' carbon content. Surprisingly, trace elements like manganese, silicon, and titanium have a lesser impact on the selection decision than assumed by the pertinent literature. Our approach can serve as an input to a range of decision-support systems, whenever a human expert needs to create groups of orders ('campaigns') that fulfil certain implicit selection criteria.**

1. Introduction

Steel is an exceptionally versatile material. Its malleability, durability, and yield strength are crucial in the construction of buildings, cars, and machines. As modern steel factories try to meet the global steel demand, they started to directly couple steelmaking, continuous casting (CC), and hot rolling to increase their economic output (Tang, Liu, Rong, & Yang, 2001). Properly planning these manufacturing steps and steadily operating the necessary equipment is imperative for the profitability of a steel enterprise because any disruptions could reduce the product quality or, even worse, force the plant to be shut down for lengthy and costly maintenance activities. In order to ensure a steady and seamless operation, an extensive number of planning constraints must be abided by (Cowling, 2003; Cowling, Ouelhadj, & Petrovic, 2004; Lee, Murthy, Haider, & Morse, 1996; Mattik, Amorim, & Günther, 2014; Park, Hong, & Chang, 2002). For instance, an important technological constraint (and simultaneously the focal point of this study) refers to the chemical compatibility of consecutively processed steel products (Tang, Wang, & Chen, 2014). The chemical compatibility of two products is basically governed by their hardness or carbon content. Expectedly, the steel hardness belongs to the most frequently discussed planning parameters (Jia, Yi, Yang, Du, & Zhu, 2013; Vanhoucke & Debels, 2009; Kosiba, Wright, & Cobbs, 1992) and is acknowledged by virtually every hot rolling planning method (Özgür, Uygun, & Hütt, 2020). Nevertheless, it is not entirely clear how steel production constraints are prioritized by practitioners and to what detail the chemical compatibility is taken into account.

Interestingly, steel production planning and scheduling continue to be heavily influenced by humans (Crawford & Wiers, 2001; Cowling, 2003; Cowling, 2001). The steel sector seems to be skeptical towards automation or digitization (Özgür, Uygun, & Hütt, 2020), although scheduling tools and software theoretically guarantee lower labor costs, improved machine utilization, increased work rates and enhanced product quality (Co, Patuwo, & Hu, 1998). For example, the selection of customer orders for immediate production is still carried out manually by human expert planners in a lot of cases. Based on their understanding of technological and commercial aspects these human expert planners pick a subset from a larger pool of customer orders and group them in production campaigns. However, some of these constraints constitute tacit knowledge because they derive from the long-term professional experience of expert planners who usually disseminate their thoughts through word-of-mouth only. This lack of documentation puts the steel plant's productivity at risk if the expert planner is absent and substituted by a less experienced worker. To the detriment of steel enterprises, planning staff shortages happen more often than not, as these expert planners are employed during normal office working hours, whereas the steel production strides ahead continuously (Cowling, 2001). Beyond that, steel manufacturers are regularly faced with situations where generating a feasible schedule appears inconceivable which compels them to ignore certain less essential constraints. Unfortunately, there is no straightforward and universally valid algorithmic solution to this problem since the exact relevance of the planning constraints commonly depends on a variety of distinct factors (e.g. product mix, type of equipment, customer requirements). As it is shown in Chapter 2 ("Literature Review"), many academic articles covering order suggestion algorithms remain extremely vague regarding the appropriate choice of selection rules and constraints – especially when it comes to the chemical compatibility of different steel grades. The existence of this research gap might further consolidate the steel industry's suspicion against automation and digitization.

In our opinion, this lack of trust can only be overcome if decision support systems comply with actual business practices in a stricter fashion. In other words, support systems need to comprehend which rules are deemed indispensable by steel decision-makers and, conversely, which constraints may be relaxed. Here, instead of creating a new algorithm or testing older ones, we concentrate on an explorative analysis of historical production data that discloses both the prioritization of the planning parameters as well as the level of complexity inherent in the decision process. To this end, we extracted association rules (with particular emphasis on chemical features) from the selection records of a human expert planner; then, we mapped these rules into a network similar to the interval graph in Lee et al. (2004). Association rules learning is a routinely used data mining method (Hahsler & Karpienko, 2017) as it represents an immensely helpful approach to discover which kinds of products (i.e. which steel grades) are preferably combined in production campaigns. Now, looking at association rules from a network / graph perspective and detecting areas of tightly connected nodes (i.e. network communities; Girwan & and Newman, 2002) allows us to identify steel grade groups that are considered linkable in the mind of the expert planner. Afterward, we compare our steel grade network to suitable randomized graphs by deploying a powerful shuffling-based network method found in Enders et al. (2018) that capitalizes on popular network concepts such as the clustering coefficient (Watts & Strogatz, 1998), and the betweenness centrality (Freeman, 1977). Finally, these methods enable the formal extraction of "first principles" that drive the selection procedure as we determine which steel grades are just connected by chance and what external steel grade attributes (e.g. chemical composition) dictate the grade togetherness. Consequently, our experiments will reveal that the human planner operates with a small set of simplistic selection rules that solely focuses on the steel's carbon content while trace elements like manganese or silicon do not play a significant role.

Previously proposed support systems (e.g. Tang, Meng, Chen, & Liu, 2016) could benefit from our work by incorporating these specific planning practices as seen in Figure 1. The remainder of this article is divided into 2. Related Work, 3. Background Information and Theory, 4. Methodology, 5. Results and Discussion, 6. Conclusion.

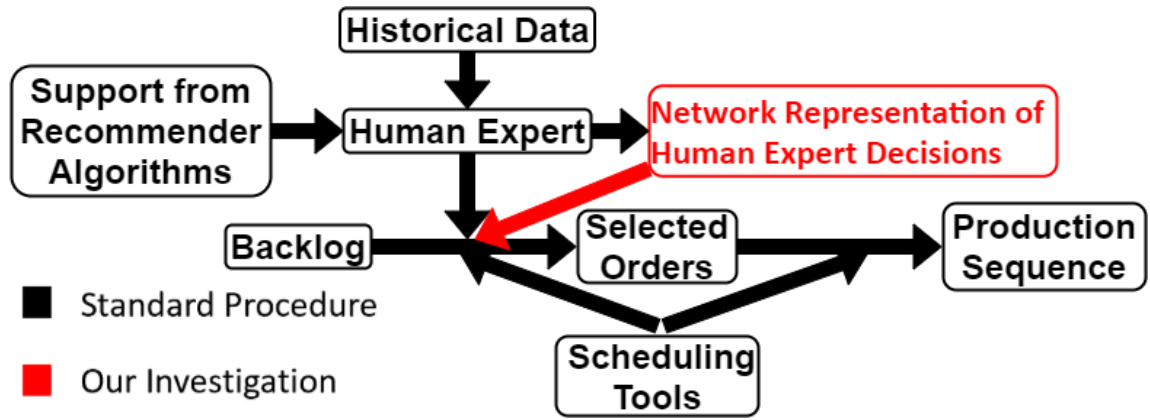

*Figure 1: Flow chart illustrating how the steel planning process could profit from our network method.*

## 2. Related Work

The following literature review is partitioned into three sub-sections that each shed light on association rule mining, complex networks, and order selection algorithms. Throughout the review, we aim at clarifying current research gaps and explaining how our article advances the state of knowledge. This will be complemented by an in-depth report on the theory / history of both association rules mining and complex networks in the third chapter ("3. Background Information and Theory").

### *2.1 Association Rules*

Association rules have been a vital tool to learn from historical data in manufacturing (Choudhary, Harding, & Tiwari, 2009; Harding, Shahbaz, & Kusiak, 2006) and other industries such as retail sales (Julander, 1992). For instance, an association rules system that facilitates production planning by mapping the relationships between products and processes was designed by Jiao et al. (2008). In addition to retail customer analysis and manufacturing in general, association rules were also adopted in steel production. Martinez-de-Pison et al. (2012) and Verma et al. (2014) mined association rules to evaluate the sources of poor corrosion protection on steel strips and the emergence of occupational accidents in a steel factory, respectively. Yet, association rules have never been applied to map the chemical compatibility of steel products into networks.

### *2.2 Complex Networks*

Note that in the subsequent chapters, we will the use terms "network" and "graph" synonymously. Li et al. (2017) have thoroughly reviewed complex networks / graphs in manufacturing. Rao (2007) illustrates how network theory concepts can enhance decision-making in miscellaneous manufacturing settings. A network-based framework to assess the complexity and affinity of products in a job shop environment is created by Jenab and Liu (2010). On top of that, a few precedent articles deal with steel scheduling by incorporating approaches from network theory (Lee, Chang, & Hong, 2004; Pacciarelli & Pranzo, 2004). Nevertheless, as far as we are aware, there is no single piece of research that merges shuffling network techniques similar to Enders et al. (2018) and data mining with the goal of exposing the implicit knowledge entrenched in the steel planning process.

### 2.3 Order Selection in Steel Manufacturing

In the past, order selection methods were adopted in the context of manufacturing as well as steel production. Cergibozan and Tasan (2019) review the literature on order picking and batching for different warehouse applications. While a few decision support systems select orders according to the monetary value (Jacobs, Wright, & Cobbs, 1988) user-assigned priority (Cowling, 2003), or simply "under some rules" (Ji & Lu, 2009), others group orders together because of their compatibility in terms of technological constraints. Here, compatibility refers to, inter alia, equal alloy composition (Balakrishnan & Geunes, 2003), similar properties (Chang, Chang, & Hong, 2000), comparable carbon contents (Tang, Wang, Liu, & Liu, 2011), suitable steel grade combinations (Tang, Wang, & Chen, 2014; Tang, Meng, Chen, & Liu, 2016) and homogenous metallurgical characteristics (Tang & Wang, 2008). By deploying these support systems, the pertinent literature promises efficient energy usage (Naphade, Wu, Storer, & Doshi, 2001; Ji & Lu, 2009), reduced costs (Tang, Yang, & Liu, 2011), optimized tundish utilization (Tang & Wang, 2008; Tang, Wang, & Chen, 2014), improved product quality (Tang & Wang, 2008), enhanced order punctuality (Naphade, Wu, Storer, & Doshi, 2001), and increased financial benefits (Tang, Meng, Chen, & Liu, 2016).

Most articles in this review do not provide information on how their algorithms select orders in detail. For example, does "suitable steel grade combinations" (Tang, Wang, & Chen, 2014; Tang, Meng, Chen, & Liu, 2016) include every chemical element that is typically comprised in steel? In fact, steel manufacturers often rely on very unique steel grade classifications aside from the publicly available industry standards. So, in some cases, a certain steel grade would conform to a precise percentage of an arbitrary element, while in others a large range of content values is implied. Analogously, "comparable carbon contents" (Tang, Wang, Liu, & Liu, 2011) could mean that only marginal carbon fluctuations are permitted within a production campaign or it could correspond to groups of steel grades that may be mutually exclusive (i.e. low carbon / high carbon). However, in order to make planning support systems implementable in real-life steel factories, it is necessary to answer these issues meticulously, or else these systems will perform worse than fully manual scheduling due to multiple reasons given in the literature: The chemical transitions between consecutively processed steel slabs are not to be modified exceedingly (Tang, Luh, Liu, & Fang, 2002; Tang & Wang, 2008), because variations of the slab hardness deteriorate the wear of the rollers (Chen, Yang, & Wu, 2008) and call for adjustments of the rolling pressure which in turn are detrimental to the steel quality

(Cowling, 2001; Cowling, 2003). Moreover, changing the chemical composition from one slab to the next prompts a transition steel grade (Tang, Wang, & Chen, 2014), which does not meet the customer requirements and consequently is of lower economic value (Lee, Murthy, Haider, & Morse, 1996; Tang & Wang, 2008; Dorn & Slany, 1994). Our experiments ("5. Results and Discussion") will try to confirm the validity of some chemical constraints based on historical production data.

## 3. Background Information and Theory

In this chapter, we summarize the theory regarding association rules mining and complex networks. The background information given here will help to understand the "methodology" and "results" sections.

### *3.1 Association Rules*

In 1992 Julander wrote up one of the first studies of association rules as a market basket analysis (Julander, 1992). By evaluating supermarket receipts he was able to infer which articles were usually bought together. Shortly after, Agrawal et al. supplied a mathematical definition of association rules along with a method that efficiently finds them − the so-called Apriori algorithm (Agrawal, Imieliński, & Swami, 1993). Because association rules might be unmanageable and untransparent, several sophisticated visualization techniques (Hahsler & Karpienko, 2017) and interest measures (Agrawal, Imieliński, & Swami, 1993; Brin, Motwani, Ullman, & Tsur, 1997; Kotsiantis & Kanellopoulos, 2006) have been established.

As mentioned above, association rules are mined through a market basket analysis. The fundamental concepts in market basket analysis are items $i_k$ and transactions $t_n = \{i_g, i_h, \ldots, i_k\}$, where a transaction consists of items that have been purchased simultaneously. In the context of steel production planning, items and transactions are congruent with customer orders (or: steel grades) and campaigns of customer orders (or: campaigns of steel grades), respectively. So, when items $i_g$ and $i_k$ are aggregated in transactions, rules in the form of $\{i_g\} \leftrightarrow \{i_k\}$ are extracted from the transaction records. To estimate the accuracy of said rules, market basket analysis resorts to statistical metrics such as support and lift. The support of an item or a group of items is the relative frequency at which the item or the group of items have been selected:

$$Support(i_k) = p(i_k)$$

where $p(i_k)$ denotes the probability that item $i_k$ is selected and

$$Support(i_g, i_k) = p(i_g, i_k)$$

where $p(i_g, i_k)$ denotes the probability that items $i_g$ and $i_k$ are selected jointly. Resting upon these relative frequencies, the lift (also referred to as "interest") formula is stated here (Brin et al. 1997):

$$Lift(i_g, i_k) = \frac{Support(i_g, i_k)}{Support(i_g) * Support(i_k)} = \frac{p(i_g, i_k)}{p(i_g) * p(i_k)}$$

As manifested by this equation, the $Lift(i_g, i_k)$ contrasts the support of an association rule against the hypothetical situation of independently occurring items $i_g$ and $i_k$. A lift value larger than one means that the corresponding items are co-selected more often than expected by chance.

*3.2 Complex Networks*

The pioneering work of Watts and Strogatz (1998) questioned the popular perception that networks are only seen in one of two very opposite forms: purely ordered or purely disordered. By steadily increasing the total disorder in an otherwise regular network through random edge rewiring, they discovered "small world" networks – a transitional type of network that distinguishes itself by a short characteristic path length similar to random graphs and, at the same time, a high clustering coefficient comparable to regular networks (Watts & Strogatz, 1998; Strogatz, 2001). Milo et al. contrasted complex networks against sets of random graphs to expose few-node sub-graphs objects (network motifs) that appear surprisingly often in complex networks (Milo, et al., 2002). Suitable random graphs were contributed by Erdős & Rényi (1960) and Molly & Reed (1995; 1998). Erdős and Rényi proposed random graphs whose node degree distribution is Poisson-distributed, whereas Molloy and Reed introduced a more adaptable type of network which is dependent on a sequence of node degrees and, hence, is able to reproduce any kind of degree distribution. An additional third kind is portrayed by random geometric graphs (Dall & Christensen, 2002). They randomly place nodes in a metric space and connect any two nodes that are not further apart than a given cut-off distance.

Girvan and Newman (2002) provided a quantitative, algorithmic framework for identifying network communities – clusters of densely connected nodes. As their community detection algorithm yields a complete dendrogram with a vast range of network divisions, it has been suggested to maximize the division's modularity. Modularity quantifies the extent to which edges occur in certain parts of the network at a higher frequency than anticipated (Newman & Girvan, 2004; Newman, 2006). Besides, useful network measures, namely the clustering coefficient $C_i$ and the betweenness centrality $b(i)$, were described by Watts & Strogatz (1998) and Freeman (1977), respectively:

$$C_i = \frac{2n}{k_i(k_i - 1)}$$

where $k_i$ / $n$ is the degree / number of nearest neighbors for node $i$ and

$$b(i) = \sum_{j \neq i \neq l} \frac{s_{jl}(i)}{s_{jl}}$$

where $s_{jl}$ and $s_{jl}(i)$ are the number of shortest paths between nodes $j$ and $l$ in total as well as the number of shortest paths between nodes $j$ and $l$ that go through node $i$, respectively.

## 4. Methods

In this section, we clarify the general methodological strategy before we shed light on the peculiarities of the data and the exact procedure of our study.

### *4.1. General Approach*

Our main goal is to deduce campaign planning rules that conjoin "first principles" known from the steel scheduling literature and actual business practices ingrained in our industrial data. Since it is usually very difficult to extract these principles from real data, a substantial part of the investigation is devoted to surveying the level of randomness in the industrial data and, thereby, the expert planner's selection behavior. Further below, it is demonstrated in detail how the combination of association rules mining and network methods facilitates the quantification of randomness. Finally, as a result of our work, we filter out important planning constraints or guidelines that could potentially be realized in the decision-support systems from section 2.3.

### *4.2. Data*

The examined production data are made available by a steel production plant where liquid steel is cast into steel slabs and rolled into coils. A sample snapshot of the steel production data is displayed in Table 1.

*Table 1: Sample snapshot of the steel production data explored in this article*

| Production Campaign | Steel Grade | Width [mm] | Thickness [mm] | Carbon Content | Manganese Content | Silicon Content |
|---|---|---|---|---|---|---|
| A | 1 | 1800 | 55 | 0.010 | 0.010 | 0.005 |
| A | 2 | 1750 | 65 | 0.010 | 0.020 | 0.015 |
| B | 3 | 1800 | 65 | 0.020 | 0.020 | 0.010 |
| B | 4 | 1750 | 70 | 0.020 | 0.010 | 0.025 |

The snapshot contains planning-relevant variables like the customer order's physical dimensions (i.e. width and thickness) and carbon content as well as labels for the respective steel grades and production campaigns. In particular, the grade and campaign labels are extremely valuable to us, as they enable the mining of association rules. While the grade labels represent the chemical ingredients of a customer order, the campaign labels summarize a set of customer orders that were manufactured consecutively. Beyond that, we append a purely theoretical column to the data: To make the temporal analysis of the planning decisions possible, the customer orders are allotted to equally-sized chronological observation windows.

### 4.3. Exact Procedure

For an illustration of the following six operational steps please refer to the supplements (see Supplements: Figure S.1). To get an overview of the data, we visually inspect the diversity of the production campaigns concerning the appearance of different steel grades. This is accomplished by means of histograms (Step 1: data visualization). Mining association rules from the production data allows us to construct a network in which the nodes symbolize the steel grades; edges between nodes are drawn for such steel grades that are frequently co-selected (Step 2: association rules mining). Within this steel grade network, we search for components that may stem from shared external attributes (e.g. chemical composition) of the intra-component nodes. Coloring the network nodes according to those attributes confirms whether the components overlap with changes of the external factors or not (Step 3: component analysis). Afterward, within these components, we look for non-random topological network fragments such as areas of high edge density (i.e. communities) and measure their relative contribution to the production output total. If any of the components exhibit intriguing community structures, we proceed to interpret them (Step 4: community analysis).

Now, we create successive evolutionary versions of the chosen network components by mining association rules in a rolling window approach with increasing window size. Correspondingly, the second evolutionary stage includes the first and second observation window, the third stage includes the first, second, and third window, etc. Then, for each node $i$ in those graphs, the degree $k_i$, the clustering coefficient $C_i$ (Watts & Strogatz, 1998), and the betweenness centrality $b(i)$ (Freeman, 1977) are calculated. Furthermore, we compute the correlations between these quantities as this provides a meaningful indication of randomness in networks (Enders, Hütt, & Jeschke, 2018). In order to estimate the extent of the detected non-random features (i.e. components / communities), we perform the same correlation evaluation on a large set of randomized graphs that resemble our steel grade network component as explained in Enders et al. (2018). With regards to this experiment, three distinct types of randomized graphs have been chosen, namely G(n,m)-random graphs according to Erdős and Rényi (1960), switch-randomized graphs (Maslov & Sneppen, 2002) as well as random geometric graphs (Dall & Christensen, 2002). The Erdős-Rényi random graphs are only required to have the same number of nodes (n) and edges (m) as the steel grade network (Erdős & Rényi, 1960); in the case of the switch randomized graphs we conserve the original degree sequence of the steel grade network (Molloy & Reed, 1995; Molloy & Reed, 1998). Conversely, the random geometric graphs only keep the number of nodes constant (Dall & Christensen, 2002). Their cut-off distance is fixed so as to approximately reproduce the total number of edges in the steel grade network. Consequently, z-scores are determined by quantifying the mean and standard deviation of these correlations for all randomized graphs; we

contrast these means against their counterpart-values retrieved from the steel grade network:

$$Z = \frac{X - \mu}{\sigma}$$

where $X$, $\mu$ and $\sigma$ are the correlation for the steel grade network, and the mean / standard deviation for all randomized graphs, respectively. Then, the entire procedure is reiterated for different evolutionary stages of the steel grade network, whereby information about the temporal development of the complexity is collected. As we aim to derive planning rules that are valid in the majority of the planning circumstances, we will first concentrate on the dominant communities. Therefore, we repeatedly prune the network by removing negligible features before we assess whether the remainder of the network is quantitatively similar to a randomized graph (Step 5: comparison with random graphs).

Finally, we attempt to draw connections between the emergence of the smaller network communities containing some rarer steel grades and the external attributes of these steel grades. Again, this is achieved by coloring the network nodes in compliance with the external parameters. Possible parameters are presented by the average chemical composition of a node or the node age; by node age, we refer to the evolutionary stage during which a given node or steel grade was introduced (Step 6: complementary analysis).

## 5. Results and Discussion

In this chapter, we will demonstrate the results obtained by carrying out the procedural steps from section 4.3.

**Step 1 (Data Visualization):** As mentioned previously, the experiment data entail, inter alia, two crucial variables: a label for each (i) steel grade and (ii) production campaign (see Table 1). Normally, one or several few steel grades are integrated by the human expert planner into every production campaign (see Figure 2 (a)). The bar plot in Figure 2 (b) highlights the dataset composition in terms of steel grades. Note the rich combinatorics arising from the number of different steel grades in a campaign and the variety of steel grades in the whole dataset. Selecting the optimal combination of steel grades might thus pose a difficult challenge for the human expert planner.

**Step 2 (Association Rules Mining):** Mining the patterns of steel grade co-selections discerned in the production data leads to the steel grade network in Figure 3 (a).

**Step 3 (Component Analysis):** Figure 3 (a) reveals the existence of two independent steel grade groups whose individual constituents are commonly selected for joint production, whereas no such links exist between these two groups. Interestingly, this separation of steel grades coincides with a transition of carbon contents (see Figure 3 (b)). Obviously, the first group comprising steel grades 11, 23, 26, and numerous less regularly fabricated steel grades has a lower mean carbon content than the second group (15, 16,

34, 44, etc.). These results align with the academic insights listed earlier, as the hardness of a steel grade – and thereby the carbon content – essentially dictates its linkability with other grades (Kosiba, Wright, & Cobbs, 1992; Chen, Yang, & Wu, 2008). We did not discover equivalent rules for other elements like manganese, silicon, or titanium (see Supplements: Figure S.2 (b) - S.2 (d)) which is unexpected since the exact contributions of these elements are considered to be important by the pertinent literature (see Section 2).

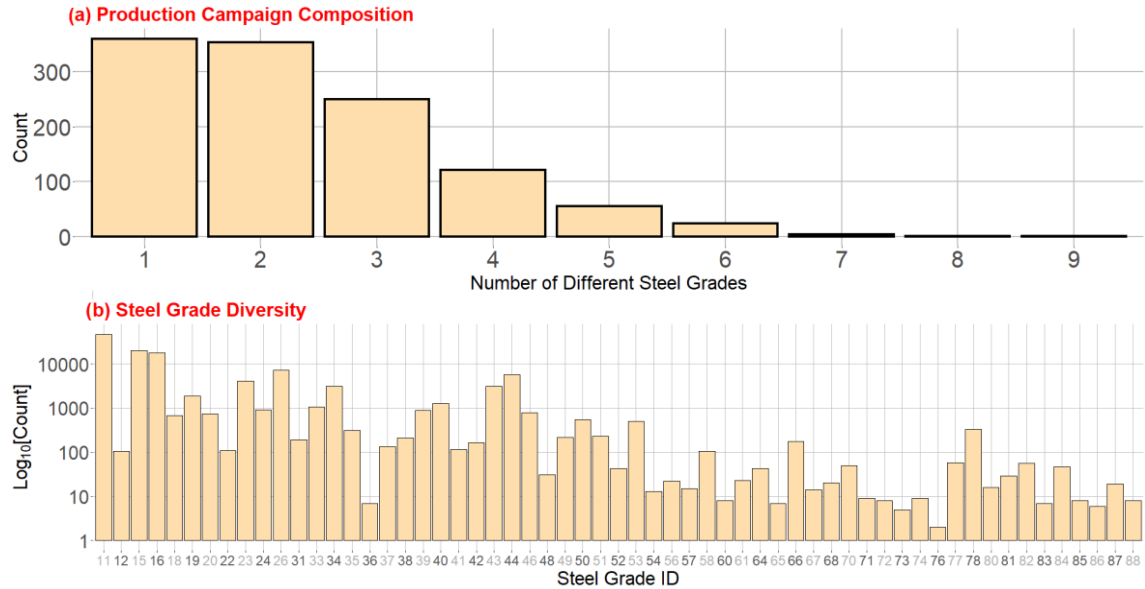

*Figure 2: (a) Histogram of the number of different steel grades per production campaign. (b) logarithmically-scaled histogram of the number of slabs per steel grade in the database.*

**Step 4 (Community Analysis):** Nevertheless, how these trace elements influence the decision process could be mirrored in the internal structure of the two carbon-related network components. That is why we break down the steel grade network from Figure 3 (a) into a number of communities and color its nodes accordingly (see Figure 4). It becomes apparent, that the network carbon component associated with lower carbon contents exhibits one massive community in the center plus three small peripheral communities. An analysis of the relative frequencies of the steel grades within each community (see Table 2) discloses that the peripheral communities contribute a negligible production output, and, hence, we can focus on the central community first. Under this assumption, we conclude that merely sorting the pending customer orders based on their carbon contents and co-selecting orders with similar carbon contents is a reasonable selection strategy regarding the mixing of steel grades.

**Step 5 (Comparison with Random Graphs):** Network communities are only one possible non-random feature of such selection strategies. Other features may still be indicative of the impact of trace elements like manganese, titanium and, silicon on the planning procedure. Therefore, we want to estimate the network's extent of randomness beyond the already confirmed carbon-related network components by studying the steel grade network as a function of increasing data volume (see Methodology: rolling window approach with increasing window size). To begin with, for successive evolutionary stages of the low-carbon network component, every node's degree, clustering coefficient, and betweenness centrality are computed and their correlations are measured.

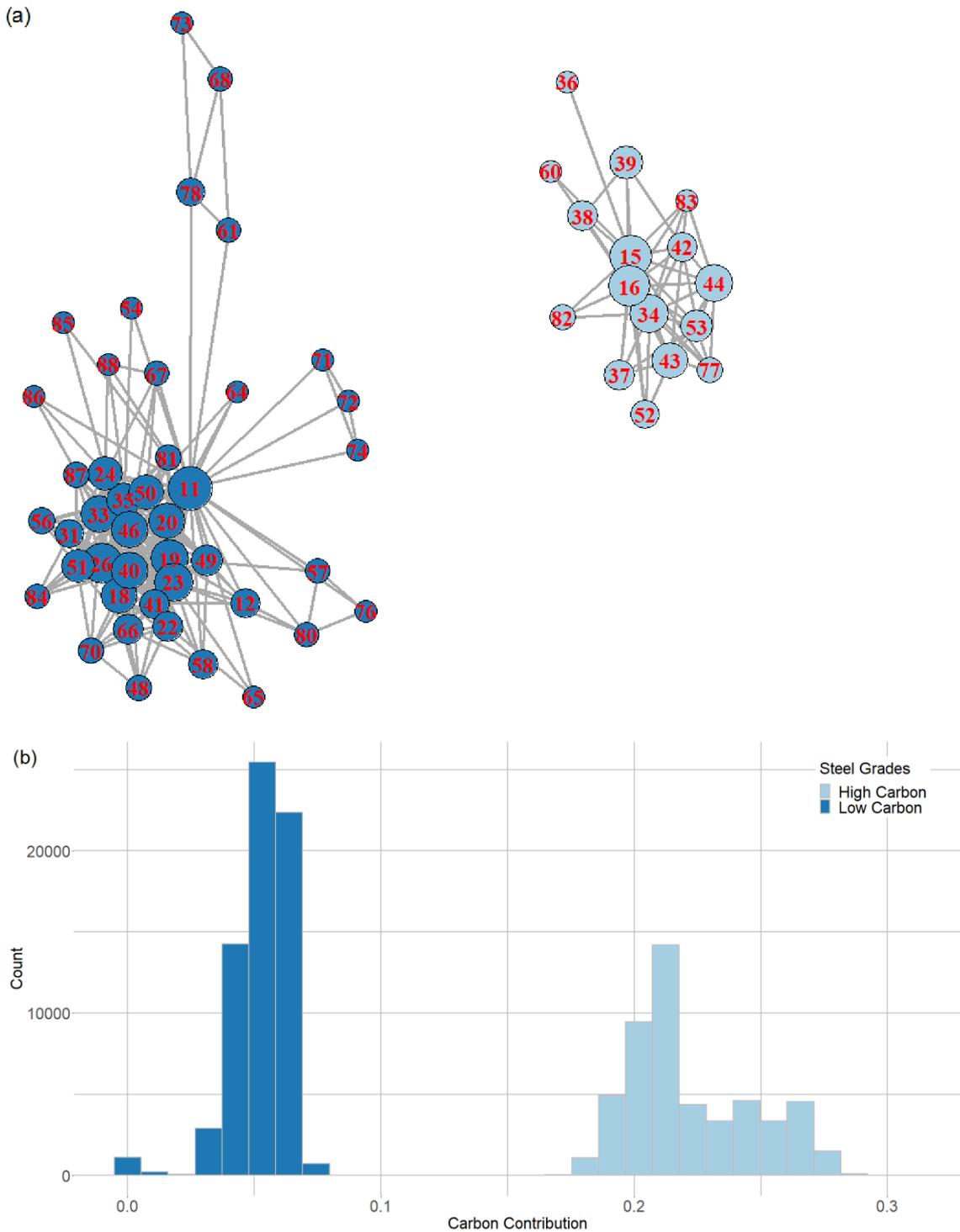

*Figure 3: (a) Steel grade network achieved through association rules mining; nodes (edges) represent steel grades (frequent joint selections of steel grades); (b) histogram of the slab carbon contribution; the colors (light and dark blue) in (b) correspond to the colors of the steel grade network in (a).*

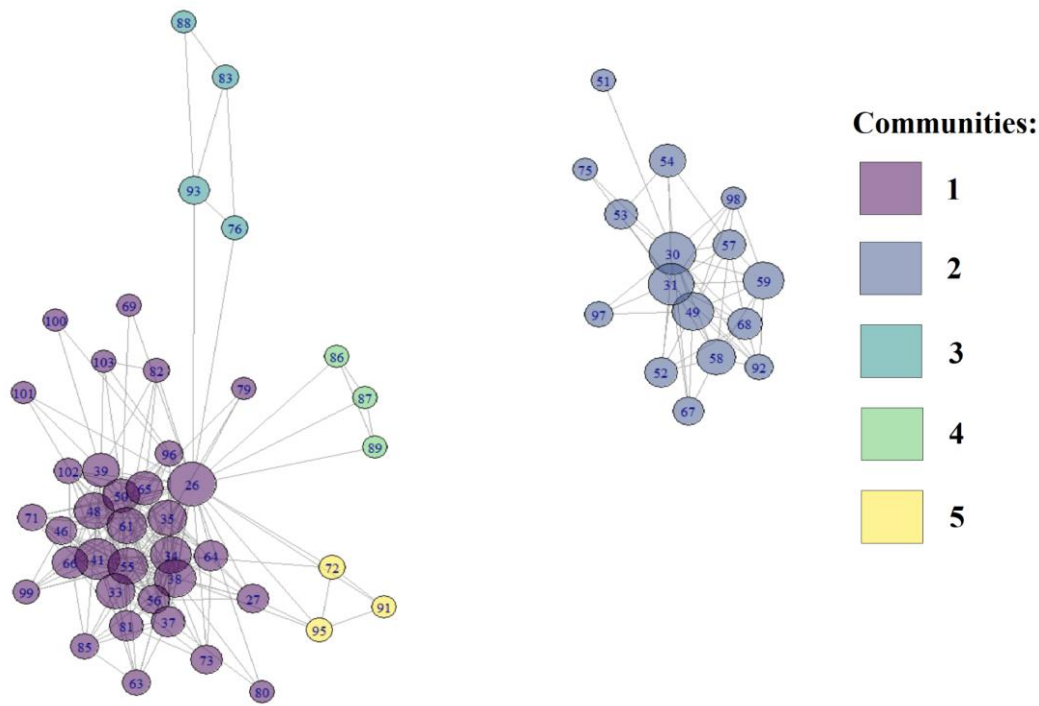

*Figure 4: Results of the community detection algorithm for the steel grade network; the communities were found by repetitively removing the edge with the highest edge betweenness and subsequently calculating the edge betweenness for the new network.*

Here, only the first and the last evolutionary stage of the steel grade network are portrayed (see Figure 5 (a) + 5 (b); note that Figure 5 b) is equivalent to Figure 4). Table 3 summarizes the correlations for the ranked distributions of degree, clustering coefficients, and betweenness centralities in both the first and the final evolutionary stage. As all four correlations are quite large, it is safe to assume that the selection of steel grades underlying the low-carbon network component involves a large amount of randomness. Whether the decrease in correlation from the first to the last evolutionary stage is accompanied by a more systematic growth of the steel grade network over time and whether this has something to do with trace elements will be addressed in the following paragraph.

Generating a substantial number of randomized graphs of three types – namely Erdős-Rényi random graphs, switch-randomized graphs as well as random geometric graphs – permits for the assessment of randomness in the low-carbon network component via z-scores. Figure 6 (a) documents the z-scores of the entire low-carbon component including all four communities for two different Spearman correlations (degree vs. betweenness centrality COR_D_BC & clustering coefficients vs. betweenness centrality COR_CC_BC) and three different types of randomized graphs (Erdős-Rényi graph ER, switch-randomized graph DEGSEQ, random geometric graph GRG) against the evolutionary stage. After the ninth evolutionary stage the z-scores for LOW_ER_COR_D_BC and LOW_DEGSEQ_COR_D_BC experience a downtrend, which does not show itself in Figure 6 (b) where we calculated the same set of z-scores, but this time we omitted the three peripheral communities. Note that we enclosed the boxplots for the Spearman z-scores of both the first and the final evolutionary stage to the supplements because they offer an intuitive visualization of how z-scores form (see Supplements: Figure S.4 (a) - S.4 (d)). The fact that the z-scores in Figure 6 (b) are much closer to zero than in Figure 6 (a) suggests that, indeed, the non-random features are a consequence of the three peripheral communities. Likewise, since all but one z-score in Figure 6 (b) persistently have a magnitude below 2 we underscore the great level of randomness in the central community which indicates a lack of universal selection rules in addition to the carbon content selection rule.

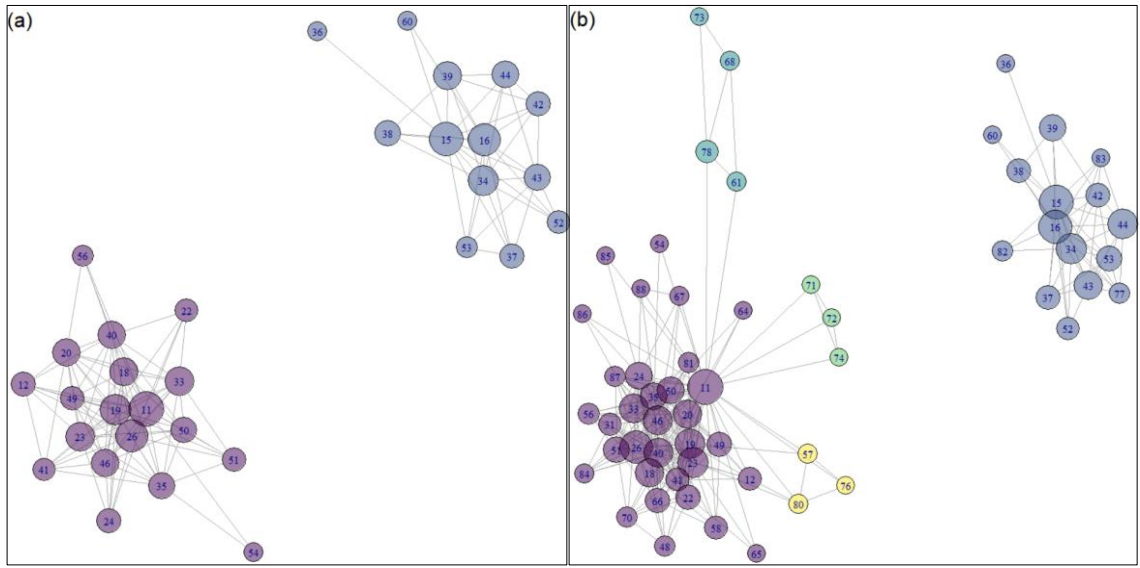

Figure 5: Results of the community detection algorithm for both the first (a) and final (b) evolutionary stage of the steel grade network; note that Figure 5 b) is equivalent to Figure 4.

Table 2: Number of different steel grades, relative frequencies of the number of slabs as well as the number of different campaigns which contain steel grades from the respective community plus the mean carbon content for each community of steel grades in Figure 4.

| Community/ Metric | Number of Steel Grades | Slab Frequency | Campaign Frequency | Mean carbon Content |
|---|---|---|---|---|
| 1 (purple) | 33 | 0.5620 | 0.5714 | 0.053 |
| 2 (blue / grey) | 16 | 0.4343 | 0.4277 | 0.222 |
| 3 (turquoise) | 4 | 0.0032 | 0.0070 | 0.004 |
| 4 (green) | 3 | 0.0002 | 0.0008 | 0.004 |
| 5 (yellow) | 3 | 0.0003 | 0.0026 | 0.056 |

**Step 6 (Complementary Analysis):** Still, it is instructive to also study the emergence of other non-random features such as the peripheral communities in the low-carbon network component. We already know that those communities are responsible for only a small fraction of the factory's production output. Moreover, many of the marginal steel grades and the peripheral communities, in particular, vary from the bulk of the steel grades concerning two external attributes: node age (i.e. during what evolutionary stage did the node occur for the first time) and node chemical composition. Figure 7 displays the same steel grade network as Figure 4, but instead it colors the nodes according to their age (dark blue - old age, light blue - young age). Evidently, the peripheral nodes were appended to the network later than most central nodes (as previously anticipated by Figure 6). Again, this supports the claim that simply complying with the carbon content constraint is already sufficient for the majority of the steel grade selection decisions. On the contrary, Figures S.2 (b), S.2 (c) and, S.2 (d) testify to the slightly dissimilar chemical compositions of the peripheral steel grades compared to the bulk, as these three figures color the network nodes in agreement with the mean concentrations of the chemical elements manganese, silicon and titanium (dark blue - low content, light blue - high content). So, although the combination strategy of steel grades within the network components is mainly not governed by the concentration of chemical elements other than carbon, these elements might however guide the steel grade selection in individual cases. Besides the prominence of the carbon content, other planning parameters such as the slab thickness and width are also recognized for their effect on selection decisions since their association rules networks have striking non-random features (see Figure S.5 (a) and S.5 (b)).

*Table 3: Spearman correlation (degree D vs betweenness centrality BC + clustering coefficient CC vs. betweenness centrality BC) and p-value for the first and final evolutionary stage of the network component associated with lower carbon contents; note that ranking the raw data rows increases the correlations between degree and betweenness centrality as well as clustering coefficient and betweenness centrality tremendously, which could imply that ranked correlation coefficients such as Kendall's Tau and Spearman's Rho may be better suited for our experiments than Pearson's correlation coefficient.*

|  | D vs. BC Start | CC. vs. BC Start | D vs. BC Final | CC vs. BC Final |
|---|---|---|---|---|
| Correlation | 0.93 | -0.99 | 0.80 | -0.95 |
| P-Value | $4.2 * 10^{-9}$ | $1.3 * 10^{-16}$ | $6.0 * 10^{-11}$ | $2.5 * 10^{-15}$ |

## 6. Conclusion

We have determined the significance of various planning parameters in the context of steel production planning by applying a mixture of association rules mining and complex network methods. To a large extent, our outcomes coincide with common knowledge about the interplay of different chemistry-related factors – with the exception that, in our dataset, elements other than carbon are largely disregarded during the manual selection process. For example, the concentration of trace elements like manganese, silicon, and titanium seems to play a tangential role in real-world factory decisions. This finding contradicts the existing literature about steel production planning constraints, as the exact chemical composition of consecutive steel grades should technically affect the steel quality (Tang, Wang, & Chen, 2014; Lee, Murthy, Haider, & Morse, 1996; Tang & Wang, 2008). Whether the lack of trace element consideration stems from the cognitive inability of the human expert planner to incorporate several planning constraints in the same instance or other commercial planning objectives (e.g. due date) are plainly more important needs to be discussed in future research.

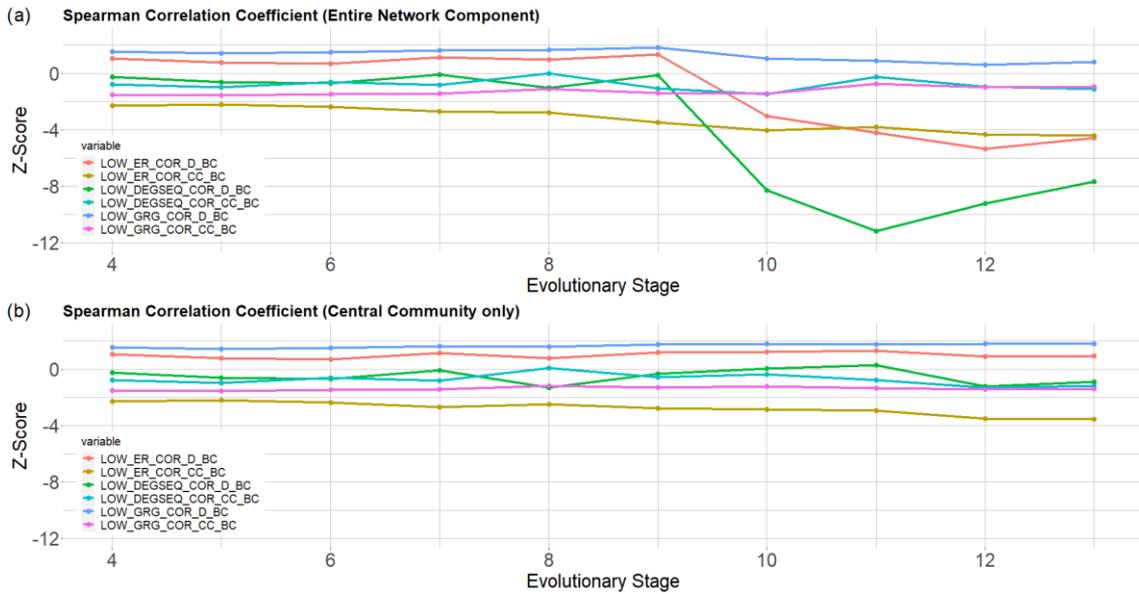

*Figure 6: Z-scores of the entire low-carbon network component (a) / the central community of said network component only (b) for two different Spearman correlations (degree vs. betweenness centrality "COR_D_BC" & clustering coefficients vs. betweenness centrality "COR_CC_BC") and three different types of randomized networks (Erdős-Rényi graph "ER", switch-randomized graph "DEGSEQ", random geometric graph "GRG") against the evolutionary stage; the corresponding Pearson and Kendall correlations are given in the supplements (see Supplements: Figure S.3 (a) – S.3 (d)).*

Our results facilitate steel production planning by explaining which planning goals or constraints cannot be relaxed if the schedule preparation appears infeasible. Consequently, if a human expert planner is absent, less experienced planners could benefit from our work and build on the expert planner's know-how. On top of that, our method helps in any situation where an order suggestion algorithm has to be configured based on implicit selection criteria that derive from fundamental technological production principles. Generally, we advocate for a stronger focus on network-aided planning and scheduling methods, since they provide an intuitive representation of the problem circumstances.

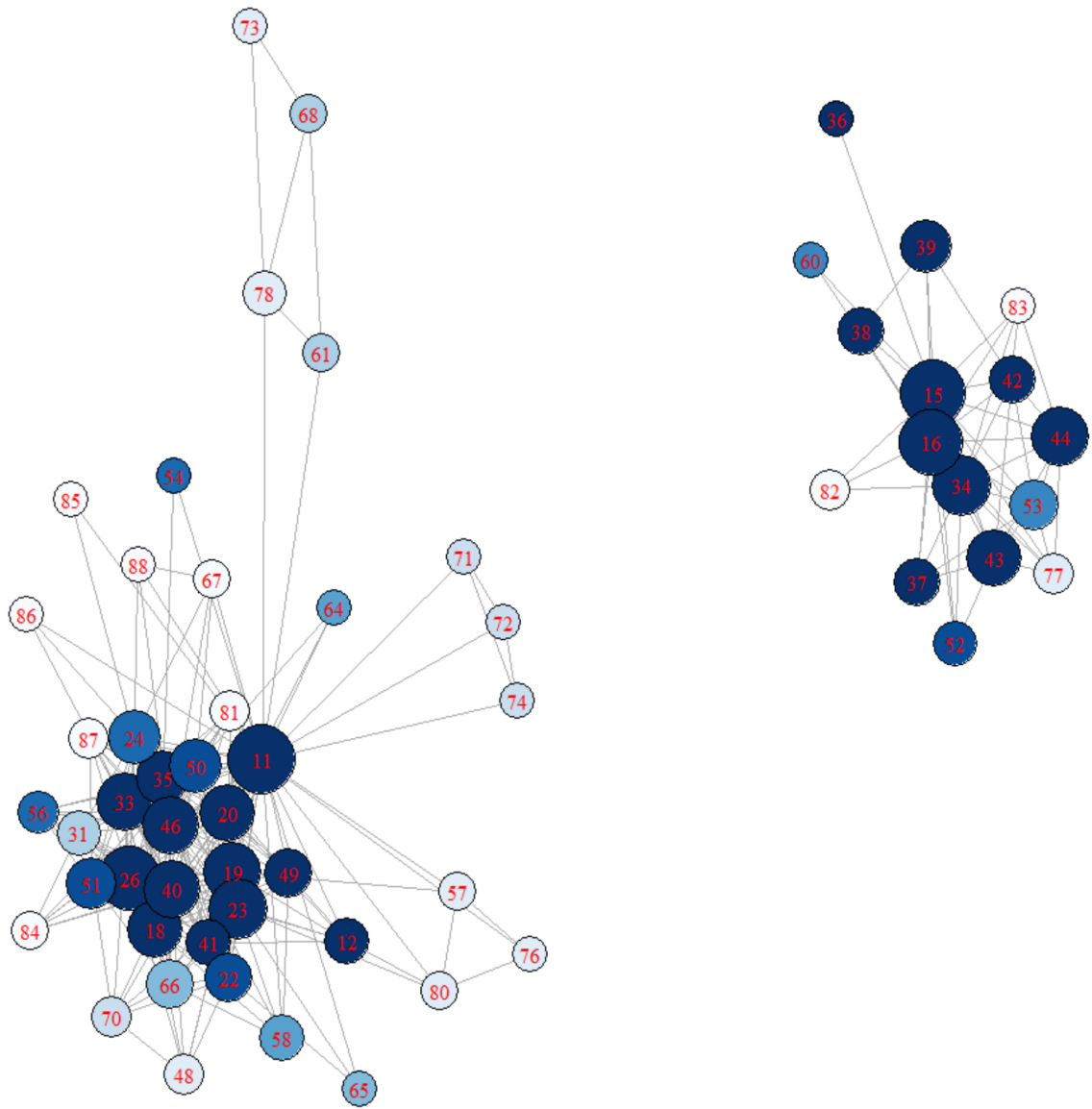

*Figure 7: Steel grade network equivalent to Figure 4 but instead the nodes are colored corresponding to the node age (i.e. evolutionary stage during which the respective steel grade was introduced; dark blue – old, light blue – young).*

# 8. Supplements

## 1. Data visualization

(a) Create steel grade histograms from production campaigns

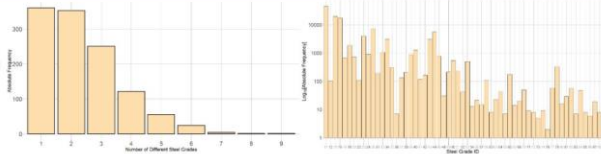

## 2. Association rules mining

(a) Find rules from production data
(b) Map rules into network

| Transaction | Items |
|---|---|
| 1 | A, B |
| 2 | A, **C, D**, E |
| 3 | B, **C, D**, F |
| 4 | A, B, **C, D** |

{**C, D**}   Frequent Itemset

{**C**} => {**D**}   Association Rule

## 3. Analysis of components

(a) Detect network components
(b) Color nodes according to external parameters (e.g. carbon content)

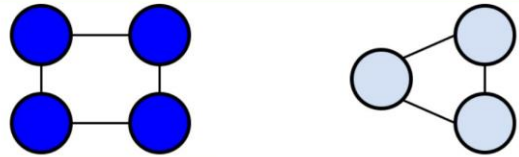

## 4. Analysis of communities

(a) Detect communities within network components
(b) Measure relative contribution of communities to production total
(c) Remove minor communities

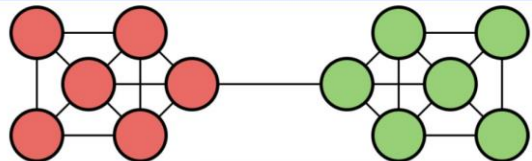

## 5. Comparison with random graphs

(a) Compute correlations between principal network measures for main communities
(b) Generate many instances of randomized graphs that resemble main communities
(c) Contrast empirical correlations against randomized ones

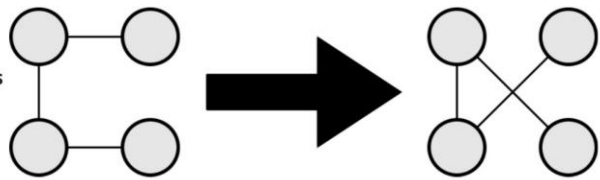

## 6. Further analysis

(a) Link minor communities to external parameters (i.e. node age)
(b) Identify additional planning-relevant factors

*Figure S.1: Step-by-step summary of our investigation*

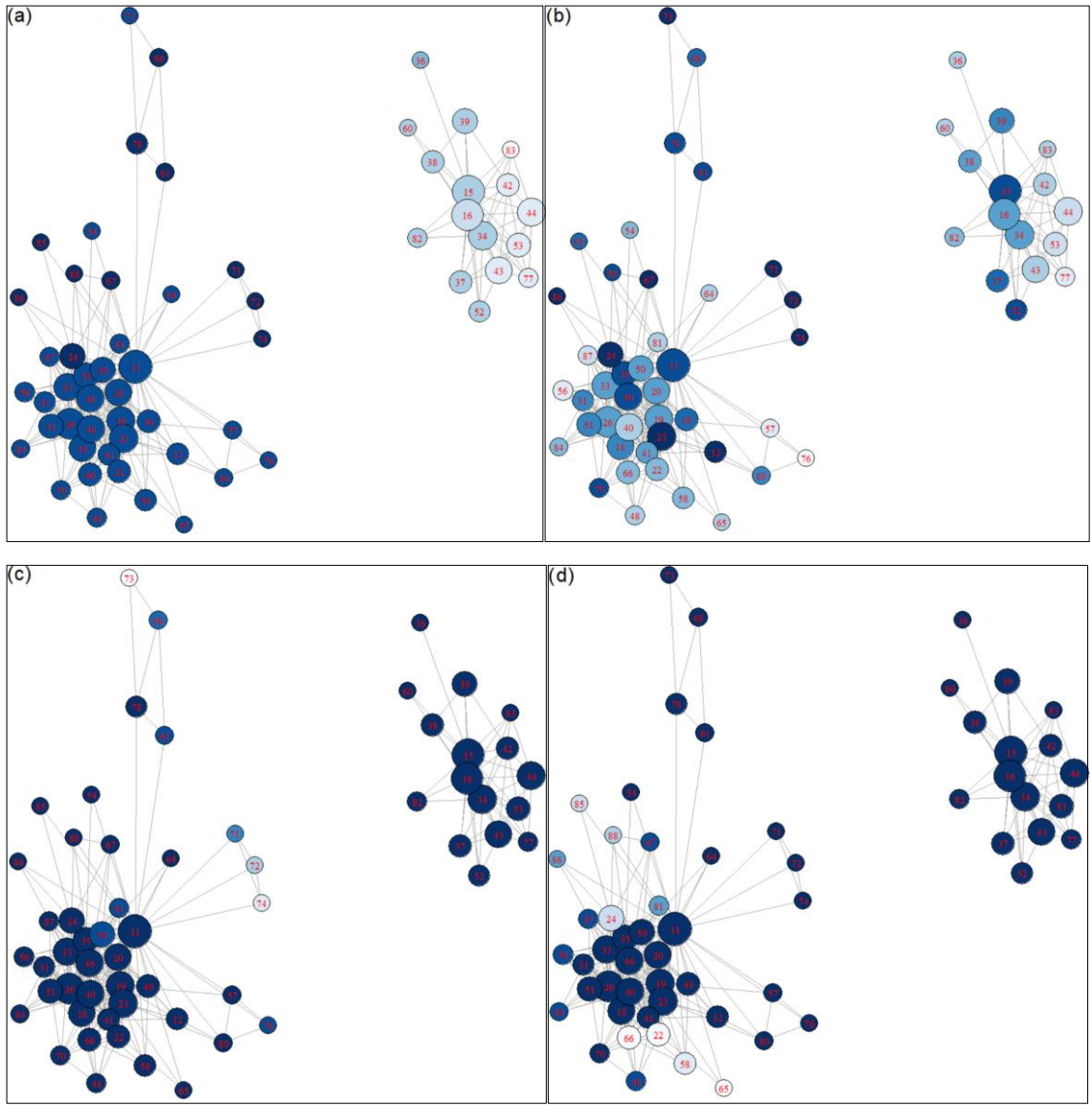

*Figure S.2: Steel grade network; the node color corresponds to the carbon (a) / manganese (b) / silicon (c) / titanium (d) content.*

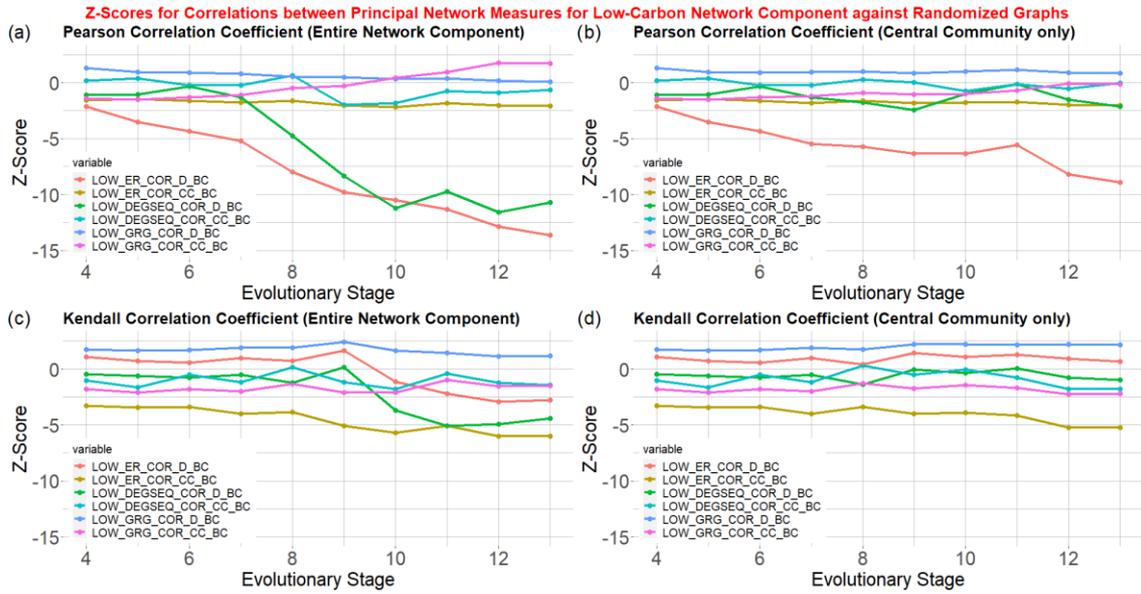

*Figure S.3: Z-scores of the entire low-carbon network component (a / c) / the central community of said network component only (b / d) for two different Pearson (a / b) and Kendall (c / d) correlations and three different types of randomized graphs (Erdős-Rényi graph "ER", switch-randomized graph "DEGSEQ", random geometric graph "GRG") against the evolutionary stage.*

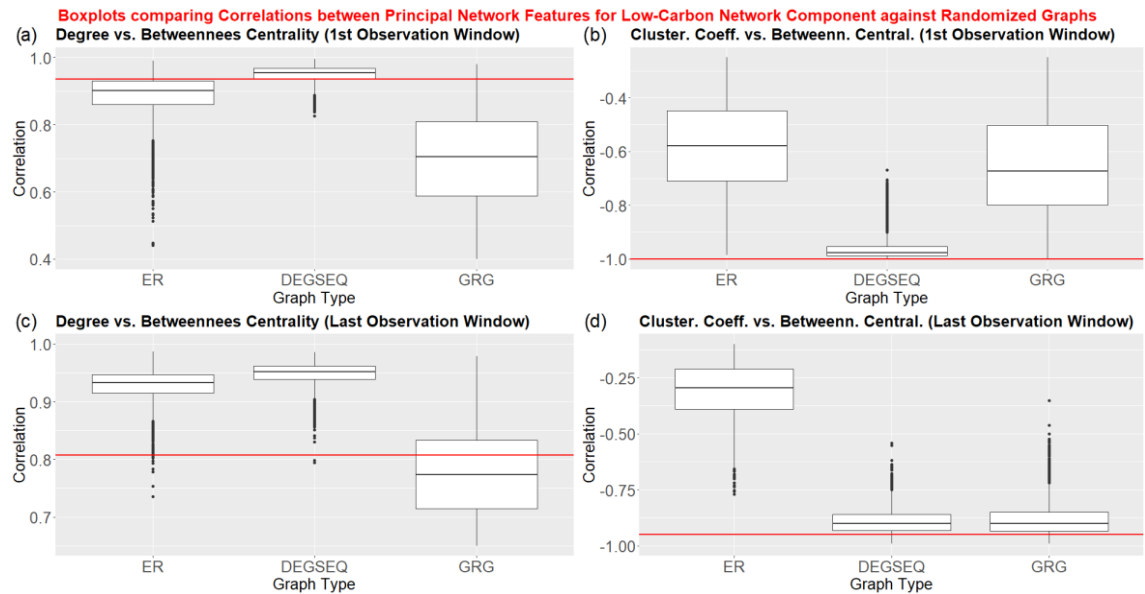

*Figure S.4: Z-scores for all four Spearman correlations mentioned in Table 3.*

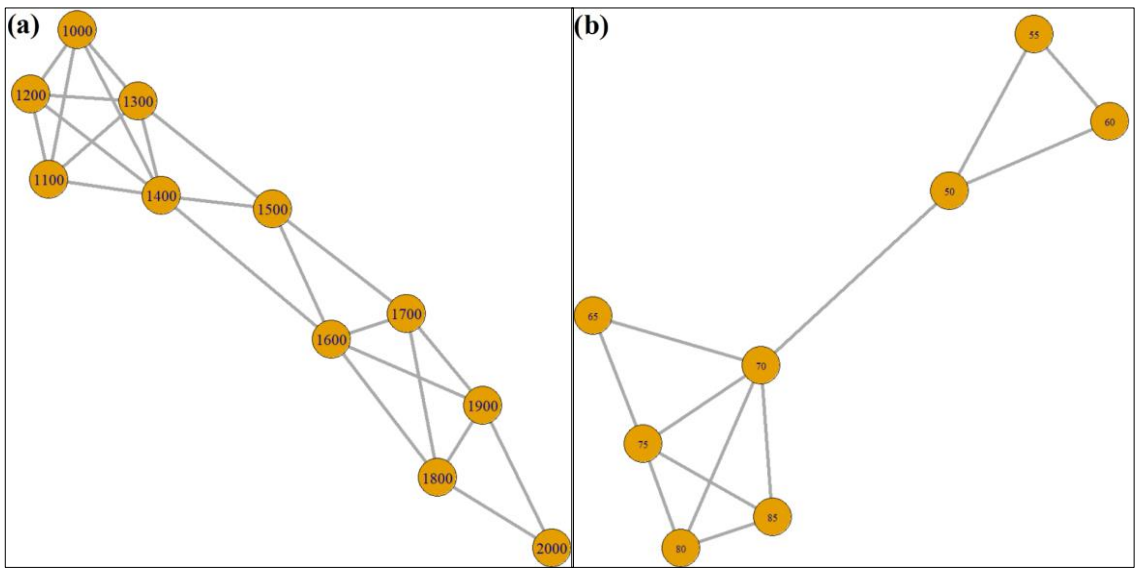

*Figure S.5: Width (a) / thickness (b) network achieved by mining association rules discerned in the production data*